\documentclass{article}

\usepackage{hyperref}
\usepackage{booktabs}
\usepackage{graphicx}

\usepackage{spconf,amsmath,graphicx}

\title{Text2Video: Text-driven Talking-head Video Synthesis with Personalized Phoneme - Pose Dictionary}
\name{Sibo Zhang, Jiahong Yuan, Miao Liao, Liangjun Zhang}
\address{
  Baidu Research, USA}
\vspace{-0.4em}
\begin{document}
%
\maketitle
\vspace{-0.5em}

\begin{abstract}
With the advance of deep learning technology, automatic video generation from audio or text has become an emerging and promising research topic. In this paper, we present a novel approach to synthesize video from the text. The method builds a phoneme-pose dictionary and trains a generative adversarial network (GAN) to generate video from interpolated phoneme poses. Compared to audio-driven video generation algorithms, our approach has a number of advantages: 1) It only needs about 1 min of the training data, which is significantly less than audio-driven approaches; 2) It is more flexible and not subject to vulnerability due to speaker variation; 3) It significantly reduces the preprocessing and training time from several days for audio-based methods to 4 hours, which is 10 times faster. We perform extensive experiments to compare the proposed method with state-of-the-art talking face generation methods on a benchmark dataset and datasets of our own. The results demonstrate the effectiveness and superiority of our approach.
\end{abstract}

\vspace{-0.4em}

\vspace{-0.4em}
\section{Introduction}
\vspace{-0.4em}



With the advance of deep learning technology, automatic video generation from audio (\textit{speech2video}) or text (\textit{text2video}) has become an emerging and promising research topic \cite{suwajanakorn2017synthesizing,thies2019neural,liao2020speech2video}. It introduces exciting opportunities for applications such as AI news broadcasts, video synthesis, and digital humans.
\textit{Speech2Video} models are trained to map from speech to video. Because there is much speaker variability in speech, \textit{Speech2Video} models need to be trained on a large amount of data, and they are not robust to different speakers. It is also less flexible to use speech as input compared to text. Furthermore, most previous methods that generate video from speech are based on LSTM to learn audio information. However, LSTM-based methods have some limitations: 1) The network needs a lot of training data. 
2) The voice of a different person degrades output motion quality. 
3) Users can not manipulate motion output such as changing speaker attitude since the network is a black box on what is learned. 
Compared to audio-based methods, text-based methods have advantages. We here define \textit{Text2Video} as a task of synthesizing talking-head video from any text input. The video generated from a text-based method should be agnostic to the voice identity of a different person.

In this paper, we propose a novel method to generate video from text. The technique builds a phoneme-pose dictionary and trains a generative adversarial network (GAN) to generate video from interpolated phoneme poses. Forced alignment is employed to extract phonemes and their timestamps from training data to build a phoneme-pose dictionary. We applied the method to both English and Mandarin Chinese. To demonstrate the effectiveness of our approach, we conducted experiments on a number of public and private datasets. Results showed that our method achieved higher overall visual quality scores compared to state-of-the-art systems. 

The main contributions of this paper are summarized as follows:
1) We propose a novel pipeline of generating talking-head speech videos from any text input, including numbers and punctuation, in both English and Mandarin Chinese. The inference time is as fast as 10 frames per second.
2) We develop an automatic pose extraction method to build a phoneme - pose dictionary from any video, online or purposely recorded. With only 44 words or 20 sentences, we can build a phoneme - pose dictionary that contains all phonemes in English. 
3) To generate natural pose sequences and videos, we introduce an interpolation and smoothness method and further utilize a GAN-based video generation network to convert sequences of poses to photo-realistic videos.

\vspace{-0.4em}
\section{Related Work}
\vspace{-0.4em}

\begin{figure}
\vspace{-0.7em}
  \centering
  \includegraphics[width=0.48\textwidth]{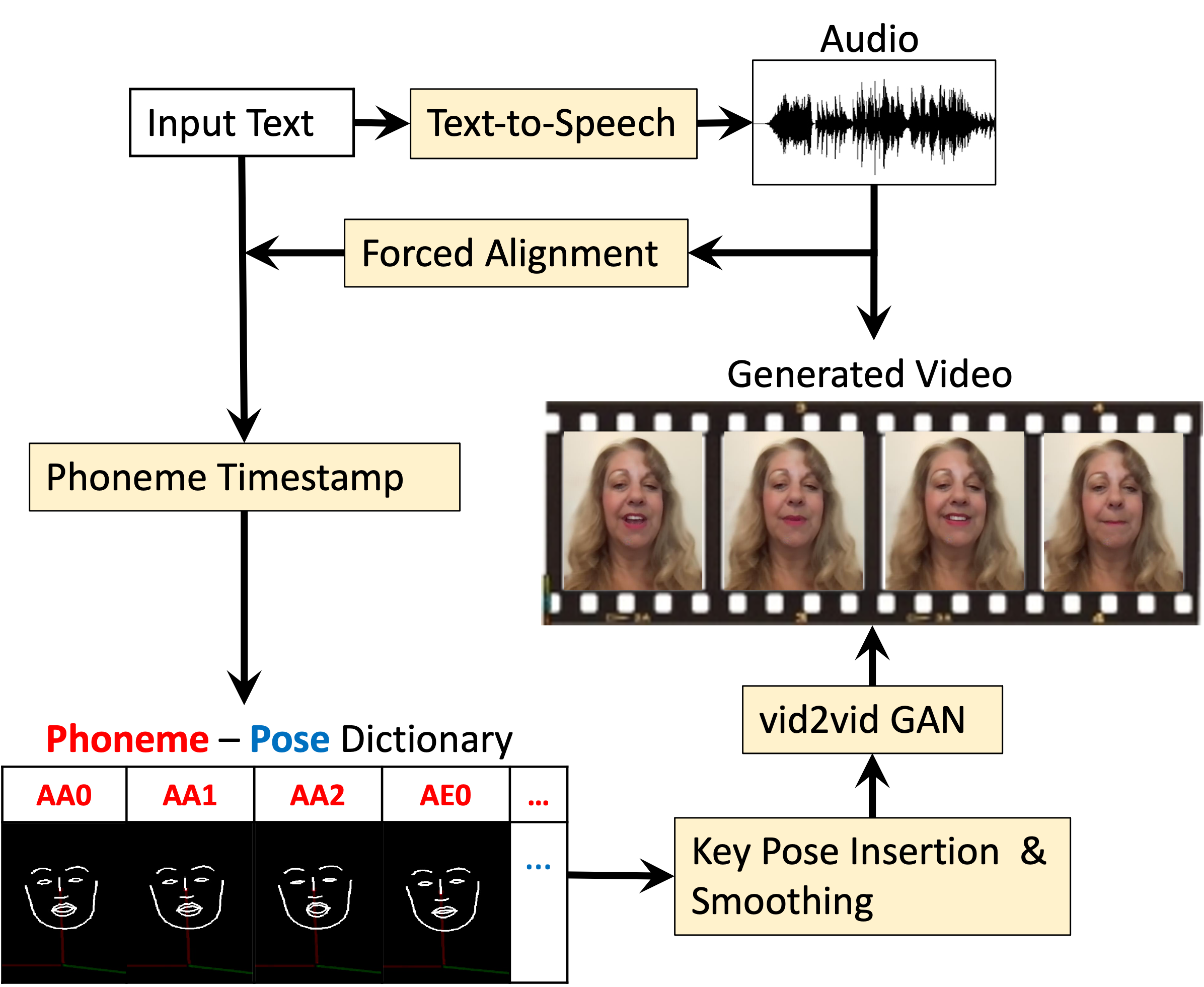}
  \caption{Pipeline of Text2Video including generating audio from text, applying forced alignment to get phoneme timestamps, searching in a phoneme-pose dictionary, applying the key pose interpolation/ smoothing module to get a sequence of poses, and generating video using modified GAN.}
  \label{fig:pipeline}
\vspace{-0.7em}
\end{figure}

\textbf{Text-Driven Video Generation.} Visual speech synthesis from text has been studied in the literature. Ezzat \cite{ezzat2000visual} introduced MikeTalk, a text-to-audiovisual speech synthesizer that converts input text into an audiovisual speech stream. Taylor~\cite{taylor2012dynamic} proposed a method for automatic redubbing of video that exploited the many-to-many mapping of phoneme sequences to lip movements modeled as dynamic visemes. 
Text-based Mouth Editing~\cite{fried2019text} is a method to overwrite an existing video with new text input. The method conducts a viseme search to find video segments with mouth movements matching the edited text. However, their synthesis approach requires a re-timed background video as input and their phoneme retrieval is agnostic to the mood in which the phoneme was spoken.

\noindent\textbf{Audio-driven video generation.} Audio-driven Video Synthesis is to drive movements of human bodies with input audio. 
For example, SythesisObama~\cite{suwajanakorn2017synthesizing} focused on synthesizing a talking-head video by driving mouth motion with speech using RNN. A mouth sequence was first generated via texture mapping and then pasted onto an existing human speech video. However, SythesisObama needs approximately 17 hours of training data for one person, which is not scalable. 
~\cite{chen2019hierarchical} utilized facial landmarks to generate video from identity image and audio signal.
~\cite{zhou2019talking} generated high-quality talking face videos using disentangled audio-visual representation.
Wang \cite{wang2020speech} proposed a GAN-based network based on the attentional multiple representations to synthesize talking-head videos from speech.
Taylor \cite{taylor2017deep} introduce a deep learning approach using sliding window regression for generating realistic speech animation. Their animation predictions are made in terms of the reference face AAM parameterization re-targeting to a character, which introduces a potential source of errors.
Ginosar \cite{ginosar2019learning} proposed a method to learn individual styles of speech gestures in two stages. However, final generated videos from their rendering stage have a few artifacts. Thies \cite{thies2019neural} developed a 3D face model by audio and rendered the output video using a technique called neural rendering ~\cite{thies2019deferred}. They proposed Audio2ExpressionNet, a temporal network architecture to map an audio stream to a 3D blend shape basis representing person-specific talking styles. 
Previously, mouth movement synthesis is mostly deterministic: given a pronunciation, the mouth's movement or shape is similar across different persons and contexts. 
Alternately, Liao \cite{liao2020speech2video} proposed a novel two-stage pipeline of generating an audio-driven virtual speaker with full-body movements. Their method was able to add personalized gestures in the speech by interpolating key poses. They also utilized 3D skeleton constraints to guarantee that the final video is physically plausible. 
However, these method are audio-based and has limitations as mentioned earlier.

\vspace{-0.4em}

\section{Method}

\vspace{-0.4em}

\begin{figure*}
   \centering
  \includegraphics[width=0.99\textwidth]{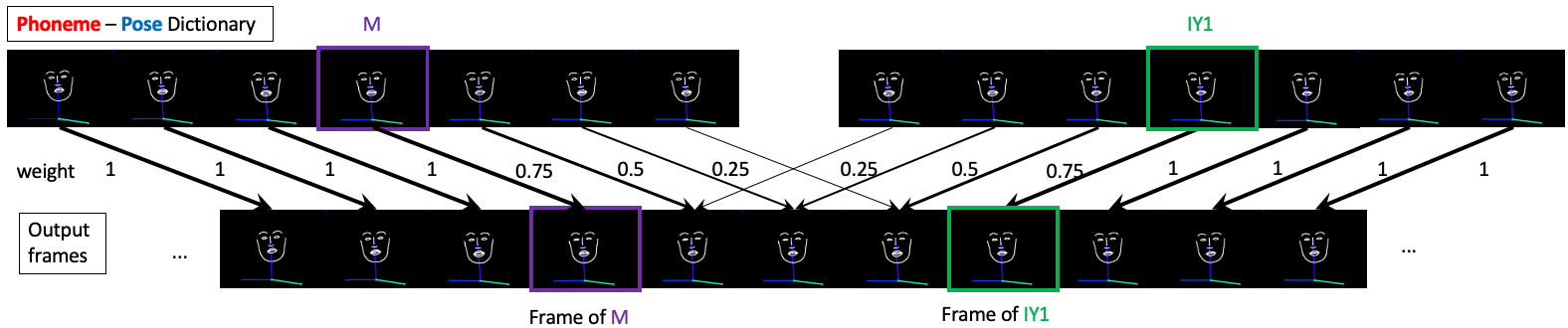}
  \vspace{-0.8em}
  \caption{
Interpolation method. To generate the output sequence of ”M IY1”, we first find the key-pose sequences of "M" and "IY" in the phoneme-pose dictionary, as well as the timestamps of the two phonemes in the output. Then we copy the two key-pose sequences to the output frames and apply interpolation to the middle frames between the two adjacent key poses.
}
  \label{fig:smooth_interpolation}
  \vspace{-0.7em}

\end{figure*}

\vspace{-0.4em}



\noindent\textbf{Text2Video Framework}. As shown in Fig.~\ref{fig:pipeline}, the input to our system is text, and the output is generated video of a talking human. Given an input text, we use TTS to generate speech from the text. Then we apply forced alignment to obtain phoneme timestamps, and lookup phoneme poses in our phoneme-pose dictionary. Next, we apply the key pose interpolation and smooth module to generate a sequence of poses. Finally, we use GAN to generate videos. 
Our method contains two key components: building a phoneme-pose dictionary from training data (audio and video of speech) and training a model to generate video from phoneme poses. 




\noindent\textbf{Phoneme-Pose Dictionary}.
Phonemes are the basic units of the sound structure of a language. They are produced with different positions of the tongue and lips, for example, with lips rounded (e.g. /u/) or spread (e.g. /i/), or wide open (e.g., /a/) or closed (e.g., /m/). English has 40 phonemes if we don't count lexical stress. 
There are three levels of lexical stress in English: primary stress, secondary stress, and unstress. Stress may influence the position of the lips in speech production. For example, the vowel 'er' in the word \textit{permit} is stressed when the word is a noun and is unstressed when it is a verb. The mouth is slightly more open when pronouncing the stressed 'er'. Therefore, we distinguish stress in the English phoneme-pose dictionary. For Mandarin Chinese, we use initials and finals as the basic units in the phoneme-pose dictionary. This is because phonemes in the finals in Chinese are more blended and don't have a clear boundary between each other \cite{ren1986}. 
We build a phoneme-pose dictionary for English and Mandarin Chinese, respectively, mapping from phonemes to lip postures extracted from a speech production video.

\noindent\textbf{Key Pose Extraction}. First, we use Openpose~\cite{cao2018openpose} to extract key poses from training videos by averaging all the phoneme-poses present in the training video. Then we build up the phoneme-pose dictionary from our phoneme extraction pipeline described below.

\noindent\textbf{Phoneme Extraction}.
We employed the P2FA aligner \cite{yuan&liberman2008} to determine phonemes and their time positions in an utterance. The task requires two inputs: audio and word transcriptions. The transcribed words are mapped into a phone sequence in advance using a pronouncing dictionary or grapheme to phoneme rules. Phone boundaries are determined by comparing the observed speech signal and pre-trained, Hidden Markov Model (HMM) based acoustic models. In forced alignment, the speech signal is analyzed as a successive set of frames (e.g., every 10 ms). The alignment of frames with phonemes is determined by finding the most likely sequence of hidden states (which are constrained by the known sequence of phonemes derived from transcription) given the observed data and the acoustic models represented by the HMMs. 
Then, we store a sequence of poses for each phoneme in the dictionary based on the alignment. The width of the phoneme-poses is determined based on the dataset video frame rate and average speaking rate.


\noindent\textbf{Text to Speech}.
We use Baidu TTS to generate audio from text input. The system's default female and male voices are used. For personalized video generation, one can use any technique to generate a voice of his/her own choice. The voice of a different person will not affect the generated video quality of our method.

\noindent\textbf{Key Pose Insertion}.
To generate a sequence of poses, we need to do key pose insertion for the missing poses between key poses. We go through all phonemes one by one in speech and find their corresponding poses in the phoneme-pose dictionary. 
When we insert a pose into a video, an interpolation is performed in their pose parameter space. We determine the interpolation strategies by taking consideration of the following factors:
phoneme poses width (which represents the number of frames for a key pose sequence extracted from the phoneme-pose dictionary), and minimum key poses distance (which determine if we need to do interpolation). Minimum key poses distance between two phonemes equals to the sum of (half of the first phoneme pose width + half of the second phoneme pose width). The equation is defined as:

\vspace{-0.4em}

\begin{equation}
    distance=\frac{1}{2} \times width_{i} + \frac{1}{2} \times width_{i+1},
\end{equation}
\vspace{-0.4em}

where $distance$ is minimum key poses distance and $width$ is phoneme pose width. 
Our interpolation strategies is:
If the interval length between two phoneme key pose frames is larger than or equal to the minimum key pose distance, we will do interpolation using the key pose\textsubscript{i} and key pose\textsubscript{i+1}. If the interval length between two phoneme key pose frames is smaller than the minimum key pose distance, we will skip the key pose\textsubscript{i+1} and using the key pose\textsubscript{i} and key pose\textsubscript{i+2} to do interpolation. 
Then, we blend key poses between two key pose sequences with a weighted sum of phoneme poses using the method showed in Fig.~\ref{fig:smooth_interpolation}. The new frames in the output sequence are interpolated between two key pose frames, weighted by their distance to those two frames. Weight is inverse proportional to the distance from a key frame which means the larger the distance, the smaller the weight.

\noindent\textbf{Smoothing}.
Smoothing is implemented after the interpolation step.   
The phoneme pose is directly copied to its time point within the video. The smoothing of the motion of poses is controlled by a smooth width parameter.
To make human motion more stable, we smooth all face keypoints except the mouth part. Because smoothing the mouth directly will sacrifice the accuracy of the mouth shape corresponding to phonemes, we calculate the mouth center and shift for all mouth key points corresponding to the center of the mouth. The new frames are linearly interpolated, weighted by their distance to other frames in the sliding window. Finally, we copy mouth key points to the mouth center of each frame. We smooth the frames one by one in the sliding window till the end of a pose sequence.  






\noindent\textbf{Training Video Generation Network}.
We utilize the generative network vid2vid ~\cite{wang2018vid2vid} to convert our pose sequences into real human speech videos. We modified the GAN network to put more weights on the face part. 

\vspace{-0.4em}
\section{Experiments}
\vspace{-0.4em}


\noindent\textbf{Datasets}. To validate our approach, we used the VidTIMIT dataset \cite{sanderson2009multi}. The VidTIMIT dataset consists of video and corresponding audio recordings of 43 people (19 female and 24 male), reading sentences chosen from the TIMIT corpus \cite{Garofolo1992}. There are ten sentences for each person. The sentences' mean duration is 4.25 seconds, or about 106 video frames (25 fps).
To test our algorithm, we also recorded a dataset of our own. We invited a female native English speaker to do recording via zoom meeting. 
We prepared prompts, including 44 words and 20 sentences. 
We also tested our algorithm in other languages like Mandarin Chinese. We used a native Mandarin Chinese speaker (female) as a model and captured a video of her reading a list of 386 syllables in Pinyin. The total recorded video is approximately 8 mins. 
Besides, we used online Youtube videos of a Chinese news broadcaster to test our algorithm. 
\noindent\textbf{Evaluation}.
To evaluate the generated videos' quality, we conducted a human subjective test on Amazon Mechanical Turk (AMT) with 401 participants. We showed a total of 5 videos to the participants in which the identities of the speakers and the video presenter are mixed. The participants were required to rate those videos' quality on a Likert scale from 1 (very bad) to 5 (very good). The ratings include 1) The face in the video is clear; 2) The face motion in the video looks natural and smooth; 3) The audio-visual alignment (lip-sync) quality; 4) The overall visual quality of the video. We compared our results with SoTA approaches, including LearningGesture~\cite{ginosar2019learning}, neural-voice-puppetry~\cite{thies2019neural}, and Speech2Video~\cite{liao2020speech2video}. Since these three methods are audio-based and use the real human voice in their demo videos. We also used a real human voice for the comparison. Table~\ref{tab:score} shows the scores from the user study for all methods. Our method has the best overall quality score compared to the other 3 SOTA methods. Besides, our text-based method is more flexible than the aforementioned audio-based method and not subject to vulnerability due to speaker variation.

\begin{table}[]
\centering
\begin{tabular}{lllll}
\hline
                         & Q1    & Q2    & Q3    & Q4    \\ \hline
LearningGesture          & 3.424 & 3.267 & 3.544 & 3.204 \\
Neural-voice-puppetry    & 3.585 & 3.521 & 3.214 & 3.465 \\
Speech2Video             & 3.513 & 3.308 & 3.094 & 3.262 \\
\textbf{Text2Video}      & \textbf{3.761} & \textbf{3.924} & \textbf{3.567} & \textbf{3.848} \\
\hline
\end{tabular}
\caption{User Study. Average scores of 401 participants on 4 questions. 
Q1: face is clear. Q2: The face motion in the video looks natural and smooth. Q3: The audio-visual alignment (lip sync) quality. Q4: Overall visual quality.
}
\label{tab:score}
\vspace{-0.3em}
\end{table}


\begin{table}[]
\centering
\begin{tabular}{@{}llllll@{}}
\toprule
                              & Q1   & Q2   & Q3   & Q4     \\ \midrule
Text2Video(w/TTS)             & 3.73 & 3.91 & 3.63 & 3.55  \\
Text2Video(w/Human voice)     & 3.78 & 4.01 & 3.71 & 3.68  \\
Real video                    & 4.02 & 4.47 & 4.46 & 4.06  \\ \bottomrule
\end{tabular}
\caption{Ablation study on different voice quality. Average scores of 401 participants on same questions as Table 2. 
}
\label{tab:video_score}
\vspace{-0.6em}
\end{table}


\begin{figure}[t]
  \centering
  \includegraphics[width=0.49 \textwidth]{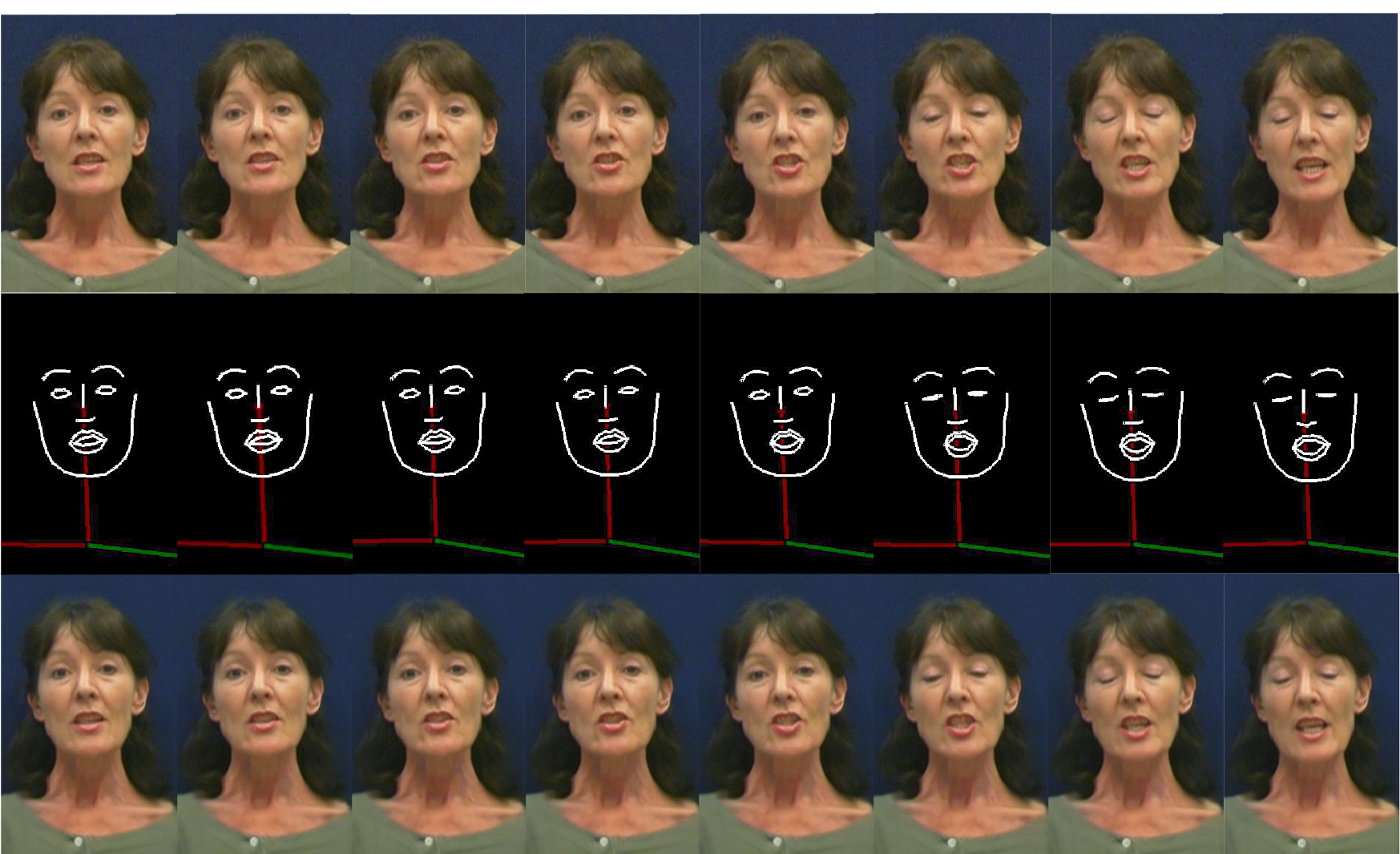}
  \caption{The output of our method from the VidTIMIT dataset. The first line shows the ground truth video clips of ``She" or ``SH IY1" in phonemes, the second line shows the output pose sequences, and the third line shows the synthesized image sequences generate from pose sequences. The result videos are at \url{https://tinyurl.com/9kfr8du2}. }
  \label{fig:VidTIMIT_result}
  \vspace{-0.6em}
\end{figure}

\noindent\textbf{Ablation Study}.
We also implemented the following user study to validate the effectiveness of our method. We showed three videos to the participants: one real and two synthesized. We used the transcription of speech in the real video to generate two synthesized videos, one with the real voice and the other with a TTS voice, to compare with the real video. We played the videos in a random order without telling the participants which one is real. As shown in Table~\ref{tab:video_score}, our output video with human voice got 3.68, and the real video got 4.06 (out of 5) on overall visual quality. The generated video is 90.6\% of the overall quality of the real video. In particular, our proposed method has similar performance on face clarity and motion smoothness compared to the real video. The video with a TTS voice got 87.4\% of the overall quality of the real video. The difference should come from the quality of the TTS audio. We simply picked an average female voice in the experiment. Using a better TTS or using a learning method to train a personalized human voice could improve the overall audio quality. Based on the user study, the overall visual quality from our text-based video generation method is barely correlated with the voice quality.






\noindent\textbf{Running Times and Hardware}.
We compare our method with SythesisObama~\cite{suwajanakorn2017synthesizing}, neural-voice-puppetry~\cite{thies2019neural}, and Speech2Video~\cite{liao2020speech2video} on training data size, data preprocessing and training time, and inference time. 
Our method needs the least amount of data to train a model. Using our fine-grained 40 words or 20 sentence list to capture all phonemes in English, the training video input can be less than 1 minute. 
Besides, our method needs the least preprocessing and training time among all four approaches. 
Preprocessing time of our approach includes running Openpose and building up a phoneme-pose dictionary. 
For the VidTIMIT dataset 
, it took about 4 hours to train our modified vid2vid GAN on Nvidia M40 GPUs while other methods need at least 30 hours. 
The inference time of our method is around 0.1 second per frame, which is similar to Neural-voice-puppetry but much faster than SythesisObama (1.5 s/frame) and Speech2Video (0.5 s/frame) on Nvidia 1080Ti. 

\vspace{-0.3em}

\vspace{-0.4em}
\section{Conclusion}
\vspace{-0.4em}

In this paper, we proposed a novel method to synthesize talking-head video from any text input. Our method includes an automatic pose extraction to build a phoneme - pose dictionary from any video. Compared to SOTA audio-driven methods, our text-based video synthesis method needs significantly less training data and has 10 times faster preprocessing and training time. We demonstrated the effectiveness of our approach for both English and Mandarin Chinese text inputs. 
\bibliographystyle{IEEEbib}
\bibliography{strings,refs}

\end{document}